\documentclass[letterpaper, 10 pt, conference]{ieeeconf}  % Comment this line out if you need a4paper

\IEEEoverridecommandlockouts                              % This command is only needed if 
\usepackage{pifont}% http://ctan.org/pkg/pifont
\usepackage{colortbl}

\title{\LARGE \bf
SuMo-SS: Submodular Optimization Sensor Scattering for \\Deploying Sensor Networks by Drones
}

\author{Komei Sugiura% <-this % stops a space
\thanks{Komei Sugiura
is with the National Institute of Information and Communications
Technology,
3-5 Hikaridai, Seika, Soraku, Kyoto 619-0289, Japan.
        {\tt\small komei.sugiura@nict.go.jp}}%
}

\usepackage{mymacros}
\usepackage{times} % assumes new font selection scheme installed
\usepackage{graphicx}

\begin{document}

\maketitle
\thispagestyle{empty}
\pagestyle{empty}

%%%%%%%%%%%%%%%%%%%%%%%%%%%%%%%%%%%%%%%%%%%%%%%%%%%%%%%%%%%%%%%%%%%%%%%%%%%%%%%%
\begin{abstract}
To meet the immediate needs of environmental monitoring or hazardous event detection, we consider the automatic deployment of a group of low-cost or disposable sensors by a drone. Introducing sensors by drones to an environment instead of humans has advantages in terms of worker safety and time requirements. In this study, we define ``sensor scattering (SS)'' as the problem of maximizing the information-theoretic gain from sensors scattered on the ground by a drone. SS is challenging due to its combinatorial explosion nature, because the number of possible combination of sensor positions increases exponentially with the increase in the number of sensors.

In this paper, we propose an online planning method called SubModular Optimization Sensor Scattering (SuMo-SS). Unlike existing methods, the proposed method can deal with uncertainty in sensor positions. It does not suffer from combinatorial explosion but obtains a $(1-1/e)$--approximation of the optimal solution. 
We built a physical drone that can scatter sensors in an indoor environment as well as a simulation environment based on the drone and the environment. In this paper, we present the theoretical background of our proposed method and its experimental validation.
\end{abstract}

% 8/8
\section{Introduction
\label{sec_intro}
}

Low-cost or disposable wireless sensors have a huge potential impact on environmental monitoring and hazardous event detection. In this study, we consider the problem of the automatic deployment of sensor networks using a drone. Typical use cases include monitoring flash floods in a desert\cite{A9}, human detection in landslides, and contamination detection on a mountain.

In most of such applications, humans are not supposed to enter the target area because of safety, cost, or other reasons; therefore, unmanned sensor deployment is required. In this paper, we use a drone to transport sensors to a target area to monitor it. Because most drones have limited battery resources, careful planning for their transportation is required to maximize a certain information-theoretic gain.

In this paper, we define a sensor scattering (SS) problem as a planning problem where drones scatter sensors in a target area to maximize a certain information criterion. In an SS problem, we have to consider the following two issues. First, because sensors are dropped from the air, their final positions on the ground are uncertain depending on the terrain and their construction material. 
\Update
Second, it is reasonable to update the plan online because of uncertainty in sensor positions.
\Done
% Second, because the maximum load of a drone is limited, it cannot carry sensors all at once. Therefore, the optimal plan cannot be specified before the task begins, but needs to be updated in an online manner.
% The drone has to carry the sensors back and forth between a sensor storage and target positions.
 % Thus, it is reasonable to store sensors in an area that humans can enter and make drones carry sensors to the target area. 

The SS problem has a close relationship with the sensor placement problem\cite{K27}. Both problems are challenging because they are typically NP-hard\cite{Y4}. The number of possible combinations of sensors increases exponentially as the number of sensors increases. Recently, Krause's work\cite{K24} proved that $(1-1/e)$--approximation can be obtained using submodularity in the mutual information criterion. This means that 63\% of the optimal score is guaranteed using a greedy method, which can avoid a combinatorial explosion. However, the method assumes that the sensor positions are known, which is invalid in SS problems.

\begin{figure}[t]
 \centering
 \includegraphics[clip,width=85mm,height=57mm]{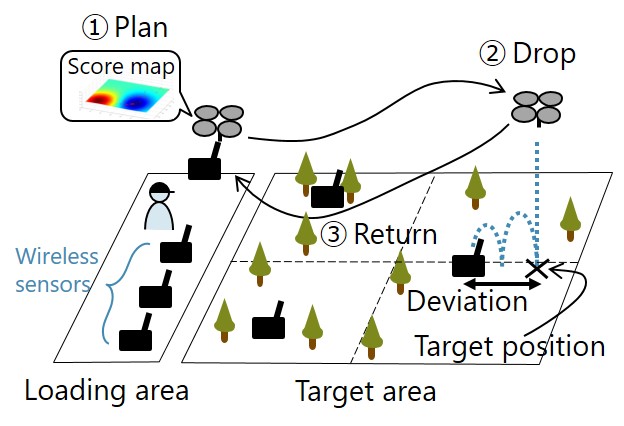}
 \caption{Typical task scenario in which the task is to deploy sensors in the target area.
 First, the drone takes off in the loading area and a sensor is attached to it.
 Given the previously scattered sensors, the next target position $\hat{y}_{pos}$ is planned by the SS method (Plan).
 The drone flies to $\hat{y}_{pos}$ and drops the sensor (Drop). The drone returns to the loading position (Return).
}
 \label{eye_catch}
\vspace{-4mm}
\end{figure}

In this paper, we propose an SS method that plans sensor positions in an online manner.
It does not suffer from combinatorial explosion but obtains a $(1-1/e)$--approximation of the optimal solution. 
A typical task scenario is illustrated in \figref{eye_catch}.
We built a customized physical drone that could scatter sensors in an indoor environment in addition to a simulation environment. In this paper, we present the theoretical background of our proposed method and its experimental validation.
\Update
To make the experimental results reproducible, the experiments were performed in the simulated environment shown in \figref{simenv}.
\Done

The following is our key contribution:
\begin{itemize}
 % \item We introduce submodular optimization\cite{N10} to drone-based sensor deployment. By doing so, we can avoid combinatorial explosion. The theoretical background is explained in \secref{sec_model}.
 \item We propose the SubModular Optimization Sensor Scattering (SuMo-SS) method \Update that considers distance-based uncertainty in sensor positions, which is relevant for practical applications. \Done The method is explained in \secref{sec_proposed}.
\end{itemize}

% Local Variables:
% eval: (auto-fill-mode -1)
% coding: utf-8
% End:

\section{Related Work
\label{related}
}

% patent Unmanned Sensor Placement In A Cluttered Terrain drop point, unmanned sensor deployment, cluttered terrain

There have been many studies on sensor placement, especially in the fields of sensor networks and robotics\cite{Y4,W9,H13,K27}. For readability, we use the term ``drone'' instead of ``Unmanned Aerial Vehicle (UAV)'' or ``multiroter helicopter''.

 % \item Some studies have investigated building sensor networks using low-cost sensors\cite{???}, robotic mapping\cite{T1,S4} and localization in sensor networks\cite{B3,H1,Wren05}.

% sensor placement
Research on optimized node placement in wireless sensor networks was previously summarized in \cite{Y4}. Some recent studies used drones for deploying sensors for optimal topology\cite{C14} or connectivity\cite{V2}. In \cite{A9}, low-cost sensors were scattered from a drone and used for detecting a flash flood; however, the work did not discuss how to optimally scatter the sensors.

In the wireless sensor network community, drone-based monitoring has been investigated to improve quality of user experience (QoE)\cite{H16}. A method to minimize a cost function based on a cover function was proposed in \cite{H16}. Energy-efficient 3D placement of a drone that maximizes the number of covered users using the minimum required transmit power was proposed in \cite{A11}.

% path planning
Uncertainty in positions, poses and maps have been widely investigated in path planning and simultaneous localization and mapping (SLAM) studies\cite{T23}. In \cite{S24}, a path planning method for mobile robots based on expected uncertainty reduction was proposed. Uncertainty in the maps and poses was modeled with a Rao-Blackwellized particle filter. Sim and Roy proposed a path planning method based on an active learning approach utilizing A-optimality\cite{S23}. In other studies, the path was planned to maximize a certain information-theoretic gain of sensors mounted on drones\cite{N11}.

The first attempt that introduced submodularity in path planning was done in \cite{C15}. Singh et al. also proposed a path planning method utilizing submodularity, and conducted real-world experiments with river- and lake-monitoring robots\cite{S22}. The submodularity objective proposed in \cite{K28} included sensor failure and a penalty reduction for the worst case. Their target application included the detection of contamination in a large water distribution network. 
% However, uncertainty in the sensor positions was not discussed.
\Update
Golovin et al. proposed the concept of adaptive submodularity in order to extend the optimization policy from a greedy method to adaptive policies\cite{G8}. In their work, uncertainty in sensor failure was discussed.
However, none of the above studies discussed uncertainty in sensor positions.
\Done

\Update
Submodularity has a close relationship with the combinatorial theory of matroids. Williams et al. recently proposed to model multi-robot tasks as functionality-requirement pairs, and applied a matroid optimization method to task allocation\cite{W8}. In their model, no uncertainty was handled. Specifically, unlike our method, their method does not consider uncertainty in sensor positions.
\Done

\Update
There have also been many attempts on alternative sensor placement methods such as evolutionary computation\cite{C16}. The method proposed in \cite{A12} can handle uncertainty in line-of-sight coverage, however it cannot handle uncertainty in sensor positions. Moreover, the method cannot be applicable to an SS problem because online planning is impossible. Indeed, most evolutionary algorithms suffer from the fact that the learning cannot be conducted in an online manner.
\Done

% SSS
Recently, drone-based monitoring has been extended to sound source localization. In \cite{W7}, a microphone array equipped to a drone was used for robustly localizing sound sources on the ground. Nakadai et al. proposed an online outdoor sound source localization method and evaluated it with a microphone array embedded on a drone\cite{N12}.

 % \item \cite{K27} explains several real-world applications utilizing submodularity including environmental monitoring using robotic sensors, activity recognition using sensors on a chair, sensor placement for contamination detection, and document selection.

% Local Variables:
% eval: (auto-fill-mode -1)
% coding: utf-8
% End:

\section{Problem Statement and Task Scenario
\label{task}
}

% \section{Task Definition
% \label{task}
% }

\begin{figure}[t]
 \centering
 \includegraphics[clip,width=87mm,height=70mm]{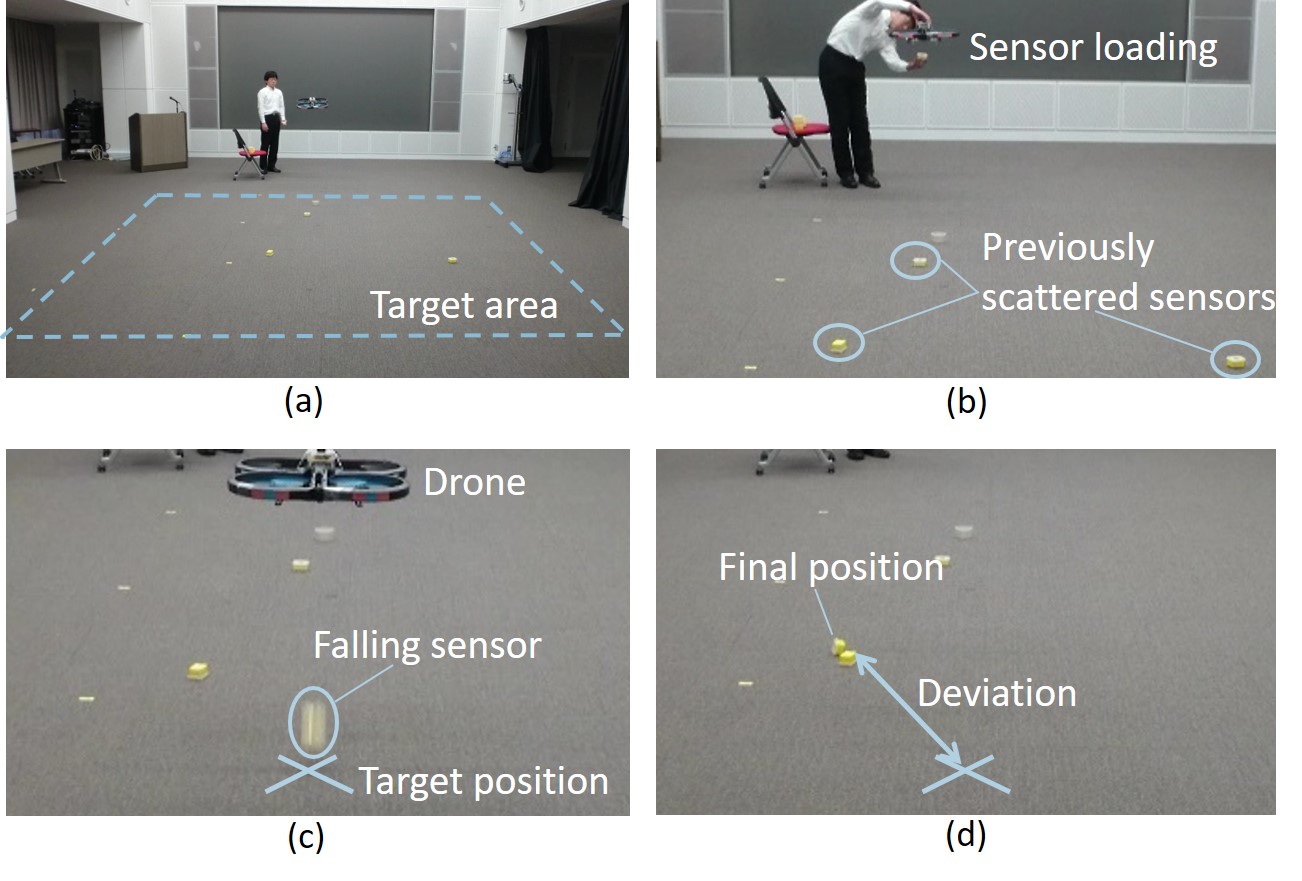}
 \caption{
 Model environment (8 $\times$ 12 m). 
 (a) Target area. (b) A sensor is attached by the experimenter. The yellow objects are sensors already scattered.
 (c) The drone drops the sensor at the target position $\hat{y}_{pos}$. (d) The sensor bounces on the terrain and stops at the final position. The double arrow represents deviation.
 }
 \label{realenv02}
\end{figure}

% \subsection{Problem Statement and Task Scenario}

In this paper, we define an SS problem as follows:
\begin{itemize}
 \item A planning problem in which \Update drones \Done scatter sensors in a target area to maximize a certain information criterion.
\end{itemize}
A typical task scenario of SS is illustrated in \figref{eye_catch}. SS is an online planning problem based on uncertain information. Previously scattered sensors affect the position of subsequent sensors, and actual sensor positions might deviate from their planned positions. In this paper, we define the term ``{\bf deviation}'' as follows:
\begin{itemize}
 \item The distance between the positions at which the drone drops the sensor and at which it lands. Although we are only considering two-dimensional deviation and distance, the method is not limited in dimensionality. In this paper, the distance is simply projected on the ground.
\end{itemize}
The following are the input and output of the planning method:
\begin{description}
 \item[{\bf Input}]: Covariance between previously scattered sensors and their target positions
 \item[\hspace{-2.5mm}{\bf Output}]: Target position of the next sensor and its informational gain
\end{description}
The target area is defined as the area to be monitored by the sensors. Humans are not supposed to enter it.

The task scenario is summarized as follows:
% The task scenario is summarized as follows\footnote{Demo video clips are available at \url{http://komeisugiura.jp/}.}:
\begin{enumerate}
 \item[(0)] Initialization: The drone takes off from the loading area. 
 % The drone goes to the sensor loading position $y_{load}$.\\
	    While the drone is hovering, the experimenter attaches a sensor to it. Although it could be autonomously loaded by the drone, that idea is outside the scope of this study.
 \item[(1)] Plan: Given the previously scattered sensors, target position $\hat{y}_{pos}$ is planned by our method. 
 \item[(2)] Drop: The drone flies to $\hat{y}_{pos}$ and drops the sensor. The actual sensor position on the ground, $y_{pos}$ is randomly deviated from $\hat{y}_{pos}$.
 \item[(3)] Return: The drone returns to the loading position, and loads the next sensor. Go to Step (1) until the maximum number of sensors have been placed.
\end{enumerate}
Demo video clip is available at this website\footnote{\url{https://youtu.be/cLx9_Zv10Oo}}.

% \begin{itemize}
%  \item We assume that the map is already built. The information necessary for 3D SLAM is already provided.
% \end{itemize}

% \subsection{Robot and Environment Models}
% removed

We assume that no remote control is performed by humans; therefore, the drone must navigate itself based on its sensor observations and a known map. Indeed in our experiments explained in Section \ref{exp}, we used a monocular SLAM method proposed by Engel {\it et al.}\cite{E2}.
% Indeed in our experiments explained in Section \ref{exp}, vision-based localization was performed based on a monocular camera.
% The localization was done by a monocular SLAM method proposed by Engel {\it et al.}\cite{E2}.
The input to the method is images taken by a monocular camera equipped with the drone. Because no external position estimation devices are used in the experiments, our method can work both indoors and outdoors.

% Local Variables:
% eval: (auto-fill-mode -1)
% coding: utf-8
% End:

\section{Sensor Model
\label{sec_model}
}

\begin{table}[t]
\caption{Symbol notations}
\centering
{\normalsize 
\begin{tabular}{c p{5cm} }
\toprule
 % \hspace{-25mm}Sensor set & \\
 $y, y'$ & Sensors\\
 $V$ & Set of target position candidates\\
 $A$ & Set of previously scattered sensors\\
 $\bar{A}$ & $V \backslash \{ A \cup y \}$\\
 $MI(A)$ & Mutual information of $A$ and $V \backslash A$\\
 $\delta_y$ & Increase in $MI(A)$ when sensor $y$ is added\\
 % \hspace{-15mm}Position and distance & \\
 $y_{pos}$ & Actual position of sensor $y$\\
 $\hat{y}_{pos}$ & Next target position\\
 $\vec{\epsilon}_{dev}$ & Deviation\\
 $\vec{\Sigma}_{dev}$ & Covariance matrix of $\vec{\epsilon}_{dev}$\\
 $d$ & Traveling distance of drone\\
 % \hspace{-25mm}Distributions \\
 $\mathcal{N}(\cdot, \cdot)$ & Gaussian distribution\\
 $K(\cdot, \cdot)$ & Kernel function\\
 $y_{obs}$ & Observation of sensor $y$\\
 $\mathcal{Y}_A$ & Observation vector of sensor set $A$\\
 $p(y_{obs})$ & Probabilistic distribution of $y_{obs}$ (Gaussian)\\
 $p(\mathcal{Y}_A)$ & Joint distribution of $\mathcal{Y}_A$ (Gaussian)\\
 $\mu_y, \sigma^2_y$ & Mean and variance of $y_{obs}$\\
 ${\bm \mu}_A, \vec{\Sigma}_{AA}$ & Mean and covariance of $\mathcal{Y}_A$\\
 $\mu_{y|A}, \sigma^2_{y|A}$ & Mean and variance of $y_{obs}$ conditioned by $\mathcal{Y}_A$\\
 $\vec{\Sigma}_{yA}$ & Covariance vector of $y_{obs}$ and $\mathcal{Y}_A$\\
 $\sigma^2_{yy'}$ & Covariance of $y_{obs}$ and $y'_{obs}$\\
\bottomrule
\end{tabular}
}
\label{tab01}
\end{table}

% \begin{table}[t]
% \caption{Symbol notations}
% \centering
% {\normalsize 
% \begin{tabular}{c p{5cm} }
% \toprule
%  % \hspace{-25mm}Sensor set & \\
%  $y, y'$ & Sensors\\
%  $V$ & Set of target position candidates\\
%  $A$ & Set of previously scattered sensors\\
%  $\bar{A}$ & $V \backslash \{ A \cup y \}$\\
%  $MI(A)$ & Mutual information of $A$ and $V \backslash A$\\
%  $\delta_y$ & Increase in $MI(A)$ when sensor $y$ is added\\
%  % \hspace{-15mm}Position and distance & \\
%  $y_{pos}$ & Actual position of sensor $y$\\
%  $\hat{y}_{pos}$ & Next target position\\
%  $\vec{\epsilon}_{dev}$ & Deviation\\
%  $\vec{\Sigma}_{dev}$ & Covariance matrix of $\vec{\epsilon}_{dev}$\\
%  $d$ & Traveling distance of drone\\
%  % \hspace{-25mm}Distributions \\
%  $\mathcal{N}(\cdot, \cdot)$ & Gaussian distribution\\
%  $K(\cdot, \cdot)$ & Kernel function\\
%  $y_{obs}$ & Observation of sensor $y$\\
%  $\mathcal{Y}_A$ & Observation vector of sensor set $A$\\
%  $p(y_{obs})$ & Probabilistic distribution of $y_{obs}$ (Gaussian)\\
%  $p(\mathcal{Y}_A)$ & Joint distribution of $\mathcal{Y}_A$ (Gaussian)\\
%  $\mu_y, \sigma^2_y$ & Mean and variance of $y_{obs}$\\
%  ${\bm \mu}_A, \vec{\Sigma}_{AA}$ & Mean and covariance of $\mathcal{Y}_A$\\
%  $\mu_{y|A}, \sigma^2_{y|A}$ & Mean and variance of $y_{obs}$ conditioned by $\mathcal{Y}_A$\\
%  $\vec{\Sigma}_{yA}$ & Covariance vector of $y_{obs}$ and $\mathcal{Y}_A$\\
%  $\sigma^2_{yy'}$ & Covariance of $y_{obs}$ and $y'_{obs}$\\
% \bottomrule
% \end{tabular}
% }
% \label{tab01}
% \end{table}

The symbol notations used in this paper are summarized in \tabref{tab01} for readability. 

First, we explain the sensor models used in this study. We assume that the sensor observations are modeled by Gaussian processes. That is, when a new sensor is introduced to the environment, its observations are modeled by a Gaussian distribution:
\begin{align}
 p(y_{obs}) &= \mathcal{N}(\mu_y, \sigma^2_y).
\end{align}
The observations obtained from the sensor set $A$ are also modeled by a Gaussian distribution:
\begin{align}
 p(\mathcal{Y}_A) &= \mathcal{N}( {\bm \mu}_A, \vec{\Sigma}_{AA} ).
\end{align}

We make the same assumption as in the previous study\cite{K24}; the covariance between two sensors can be approximated by a radial basis function (RBF) kernel using sensor positions as its parameters.
Thus, the covariance between sensors $y$ and $y'$ is modeled as follows:
\begin{align}
 % \sigma^2_{yy'} \simeq K(y_{pos}, y'_{pos}) &= \exp \left\{ \frac{||y_{pos} - y'_{pos}||^2}{2 \gamma^2}
 \sigma^2_{yy'} \simeq K(y_{pos}, y'_{pos}) &= \exp \left\{ \frac{||y_{pos} - y'_{pos}||^2}{2 \phi^2}
 \right\},
\end{align}
where $\phi$ denotes the kernel's parameter. The intention of the above equation is that close sensors will have similar values.

We assume that a sensor is dropped at one of the target candidates defined in the target area beforehand. Let $V$ and $A$ be a set of the target candidates and a set of previously selected target positions, respectively. Krause's method\cite{K24} uses mutual information as information gain by introducing a new sensor $y$ given $A$. Let $MI(A)$ be the mutual information between observations obtained from $A$ and $V \backslash A$:
\begin{align}
 MI(A) \triangleq I(A;V \backslash A).
\end{align}
Note that we cannot directly obtain observations from $V \backslash A$; therefore, we use the sensor model.

When a sensor $y$ is newly introduced, the increase in $MI(A)$ is:
\begin{align}
 \delta_y = MI(A\cup y) - MI(A).
\end{align}
Although a greedy method does not always give the optimal solution in general, it is guaranteed to give $(1-1/e)$--approximation for monotonic submodular functions\cite{N10}. $MI(A)$ is a monotonic submodular function when the number of sensors is less than $|V|/2$\cite{K24}.
Because $(1-1/e)$ is approximately 63\%, this means that 63\% of the optimal score is guaranteed even in the worst case. In a typical sensor placement task, 90\% of the optimal score is empirically reported in the above work.

Under a condition where sensors can be placed without uncertainty, the near-optimal target position $\hat{y}_{pos}$ is obtained as follows:
\begin{align}
 \hat{y}_{pos} &= \argmax_{y \in V\backslash A} \delta_y.
\label{eq41}
\end{align}
Details are explained in Appendix B.

\section{Proposed Method: SuMo-SS
\label{sec_proposed}
}

The main difference between the ordinary sensor placement problems and SS is that sensor positions have uncertainty. Instead of \eqref{eq41}, SuMo-SS maximizes the expectation of $\delta_y$ over a deviation distribution as follows:
\begin{align}
 \hat{y}_{pos} &= \argmax_{y \in V\backslash A} 
\left\{ \mathbb{E}_{pos}[MI(A\cup y)] - \mathbb{E}_{pos}[MI(A)]\right\}\nonumber\\
 &= \argmax_{y \in V\backslash A} 
\mathbb{E}_{pos}[MI(A\cup y) - MI(A)].
%  &= \argmax_{y \in V\backslash A} 
% \mathbb{E}_{pos}
% \left[
% \frac{ \sigma^2_y - \vec{\Sigma}_{yA} \vec{\Sigma}_{AA}^{-1} \vec{\Sigma}_{Ay} } 
% {\sigma^2_y - \vec{\Sigma}_{y\bar{A}} \vec{\Sigma}_{\bar{A}\bar{A}}^{-1} \vec{\Sigma}_{\bar{A}y}}
% \right]
\label{eq43}
\end{align}
\Update
In Appendix A, we explain that the above expected mutual information is submodular.
\Done
Using the transformation explained in Appendix B, we obtain the following:
\begin{align}
 \hat{y}_{pos}&= \argmax_{y \in V\backslash A} 
\mathbb{E}_{pos}
\left[
\frac{ \sigma^2_y - \vec{\Sigma}_{yA} \vec{\Sigma}_{AA}^{-1} \vec{\Sigma}_{Ay} } 
{\sigma^2_y - \vec{\Sigma}_{y\bar{A}} \vec{\Sigma}_{\bar{A}\bar{A}}^{-1} \vec{\Sigma}_{\bar{A}y}}
\right].
\label{eq44}
\end{align}
To obtain the expectation above, we model the final position of a dropped sensor as follows:
\begin{align}
 y_{pos} &= \hat{y}_{pos} + \vec{\epsilon}_{dev}\\
 \vec{\epsilon}_{dev} &\sim \mathcal{N}( \vec{0}, \vec{\Sigma}_{dev} ).
\end{align}
This means that the deviations are modeled by a Gaussian distribution, where the mean is zero and the covariance matrix is $\vec{\Sigma}_{dev}$. Because prior knowledge about the sensor materials and the environment's terrain is given in most practical applications in industry, we assume that $\vec{\Sigma}_{dev}$ is set with reasonable values by the developer.

In the preliminary investigation with the physical environment shown in \figref{realenv02}, the deviation was mainly dependent on the distance from the loading position and the directions ($x$ and $y$ axes); therefore, we model $\vec{\Sigma}_{dev}$ as a linear combination as follows:
\begin{align}
 \vec{\Sigma}_{dev} &=
\begin{bmatrix}
 w_1 d + \gamma & \gamma\\
 \gamma & w_2 d + \gamma
\end{bmatrix}
,
\end{align}
where  $d$ denotes the Euclidean distance between the loading position and $\hat{y}_{pos}$, $(w_1, w_2)$ denotes weight parameters with regard to the directions, and $\gamma$ denotes a positive small number so that the variances are always strictly positive.
% $($ d = ||y_{init} - y_{target}^{*})||$, 

Although the distribution of the previously scattered sensors should be continuous, the expectation can be approximated by a discrete mesh with appropriate granularity. \eqref{eq44} can be computed in parallel because such discrete points are independent.

% \begin{algorithm}
%     \caption{Euclid's algorithm}
%     \label{euclid}
%     \begin{algorithmic} % The number tells where the line numbering should start
%         \Procedure{Euclid}{$a,b$} \Comment{The g.c.d. of a and b}
%             \State $r\gets a \bmod b$
%             \While{$r\not=0$} \Comment{We have the answer if r is 0}
%                 \State $a \gets b$
%                 \State $b \gets r$
%                 \State $r \gets a \bmod b$
%             \EndWhile\label{euclidendwhile}
%             \State \textbf{return} $b$\Comment{The gcd is b}
%         \EndProcedure
%     \end{algorithmic}
% \end{algorithm}

% Local Variables:
% eval: (auto-fill-mode -1)
% coding: utf-8
% End:

\section{Experiments
\label{exp}
}

\Update
To validate our method, we conducted simulation experiments in which SuMo-SS was compared with a reasonable baseline method. In the following, we first explain the physical drone and environment that were used for building the simulation. Then, we explain qualitative and quantitative results.
\Done

\subsection{Robot and Environment Models}

The model environment (8 $\times$ 12 m) in this study is shown in \figref{realenv02}. We assume that its map is already known. Because outdoor environments have many uncontrollable effects on sensors and actuators, we assume an indoor environment. This does not mean that the proposed method is limited to indoor environments.

To build the drone used in the experiments, the following specifications should be addressed:
\begin{itemize}
 \item[(a)] It must have a mechanism for attaching/detaching a sensor.
 \item[(b)] Its propelling power must be adequate to carry a load of at least one sensor.
 \item[(c)] It must be sufficiently small to conduct experiments under controlled indoor environments.
 % \item[(d)] Its main hardware components should be easily available, and its software component should be compatible with such standard middleware as Robot Operating System (ROS).
 \item[(d)] Its main hardware components should be easily available, \Update and its software component should be based on standard libraries so that the experimental results can be easily reproduced. \Done
\end{itemize}

% Most commercial drones do not satisfy the above item (a); therefore, we customized a base platform that is commercially available. As the based platform, we selected the Parrot AR.Drone 2.0 that has a ROS-Gazebo compatible simulation model. The robot platform is shown in \figref{drone}. Item (d) is important because most commercial drones have low-cost parts whose characteristics deteriorate over time, which makes it difficult to use the same hardware for long periods.

\begin{figure}[b]
\centering
\begin{minipage}[c]{45mm}
 \includegraphics[bb=0 0 816 612,height=35mm]{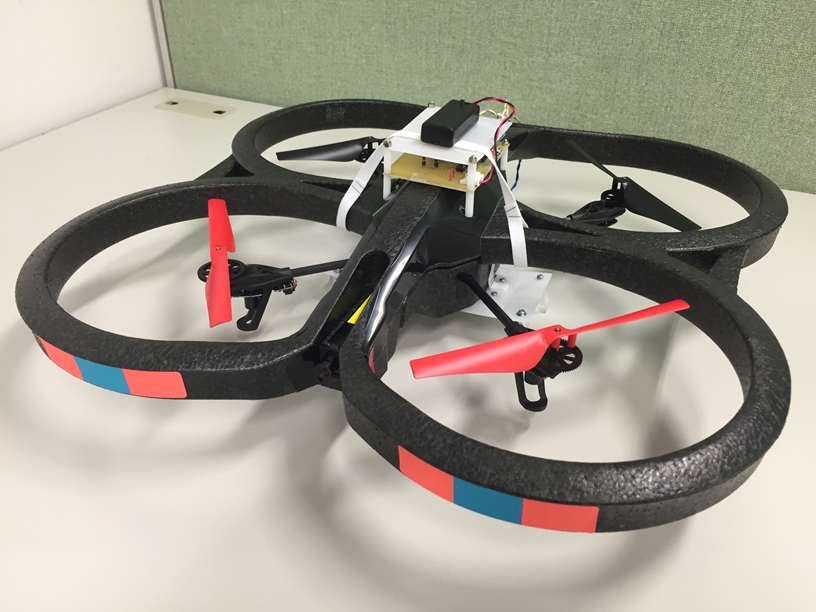}
\end{minipage}
\hspace{1mm}
\begin{minipage}[c]{35mm}
 \includegraphics[bb=0 0 385 357,height=35mm]{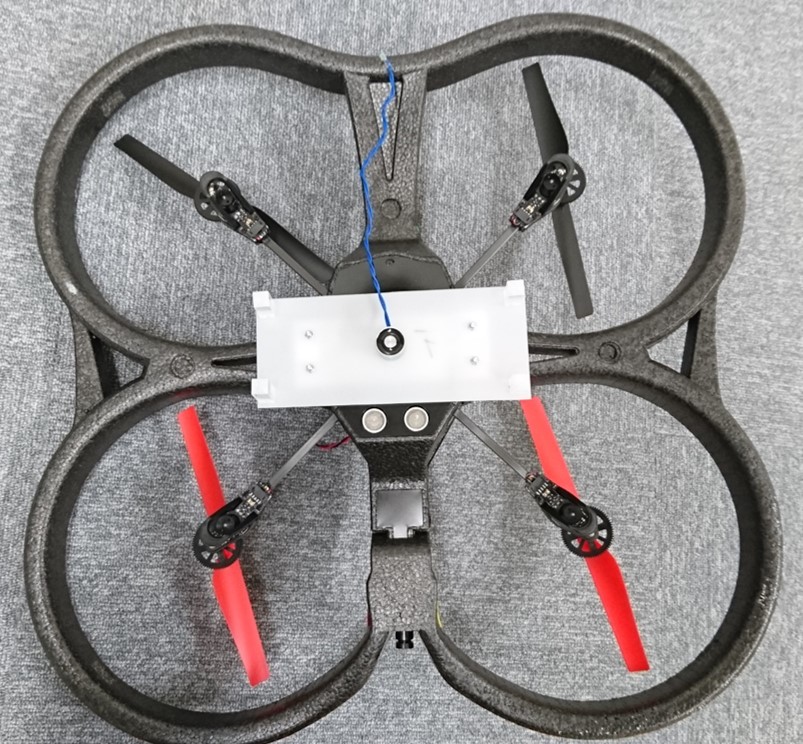}
\end{minipage}
 \caption{Physical platform used in this study. Left: Top view. Right: Bottom view. It has an electromagnetic device for attaching/detaching a sensor in the center.}
 \label{drone}
\end{figure}

Most commercial drones do not satisfy the above item (a); therefore, we customized a base platform that is commercially available. As the base platform, we selected the Parrot AR.Drone 2.0 that has a ROS-Gazebo compatible simulation model. The robot platform is shown in \figref{drone}. Item (d) is important because most commercial drones have low-cost parts whose characteristics deteriorate over time, which prevents us from conducting experiments with the same hardware for long periods.

We built a mechanism for attaching/detaching a sensor for the drone. The mechanism consists of an electromagnetic device controlled by a newly developed ROS module. The device can attach/detach a sensor when the electromagnetic power is turned on/off. The maximum load is approximately 50 g, under conditions in which the drone can fly stably. In this study, we assume that the sensor is light-weight (50 g or less) and has a metal part that can be attached to the drone by electromagnetic force. We also assume that each sensor is manually attached to the drone individually, which means that the sensors are not autonomously loaded. Although we assume that the drone can carry one sensor at a time, the method is not limited to this number of sensors.

To make the experimental results reproducible, a simulation environment shown in \figref{simenv} was used in this study. For this purpose, the above hardware was modeled as a simulated robot. We used Gazebo for the simulator and ROS for controlling the drone in the experiments.

% Local Variables:
% eval: (auto-fill-mode -1)
% coding: utf-8
% End:

\subsection{Experimental Settings}

\begin{figure}[b]
 \centering
 \includegraphics[clip,width=85mm,height=60mm]{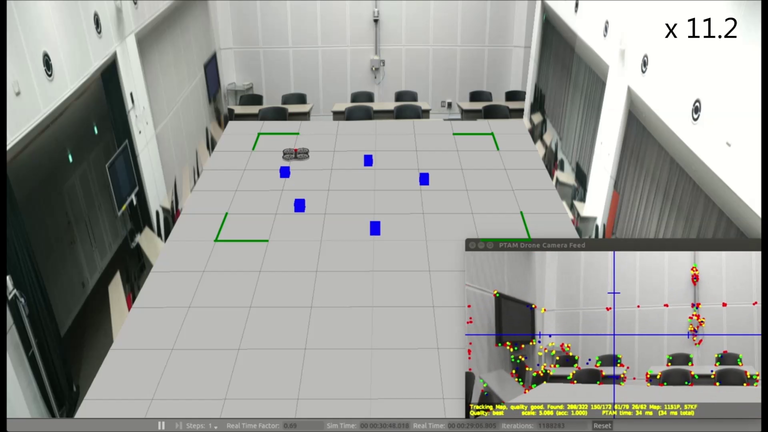}
 \caption{
 Simulation environment used in the experiments. The right bottom image is a sample camera image taken by the drone.
 The blue cubes represent sensors.
 The sensors are supposed to be placed in the area inside the green lines.
 }
 \label{simenv}
\end{figure}

Experiments were conducted in the simulation environment shown in \figref{simenv}. In the figure, the blue cubes represent the deployed sensors. In the environment, we set 25 grid points as the target candidates, $V$. The 25 grid points were equally positioned within the \Update 5$\times$5 \Done m area surrounded by the green lines. Note that even if a sensor was dropped inside the area, it might land outside of it because of deviation.
% Experiments were carried out in the simulation environment shown in \figref{simenv}. In the figure, the blue cubes represent the deployed sensors. In the environment, we set 25 grid points as the target candidates, $V$. The 25 grid points were equally positioned within the 4$\times$4 m area shown by the green lines. Note that even if a sensor is dropped inside the area, it might land outside of it because of deviation.

A sample camera image taken by the drone is shown in the bottom-right panel of \figref{simenv}. Feature points detected by the aforementioned monocular SLAM method\cite{E2} are shown as red, green, and yellow dots.

\subsection{Qualitative Results}

\begin{figure}[tpb]
 % \centering
%% 2
\begin{minipage}[c]{38mm}
 \centering
 \includegraphics[clip,trim=80 70 100 0,height=40mm,width=38mm]{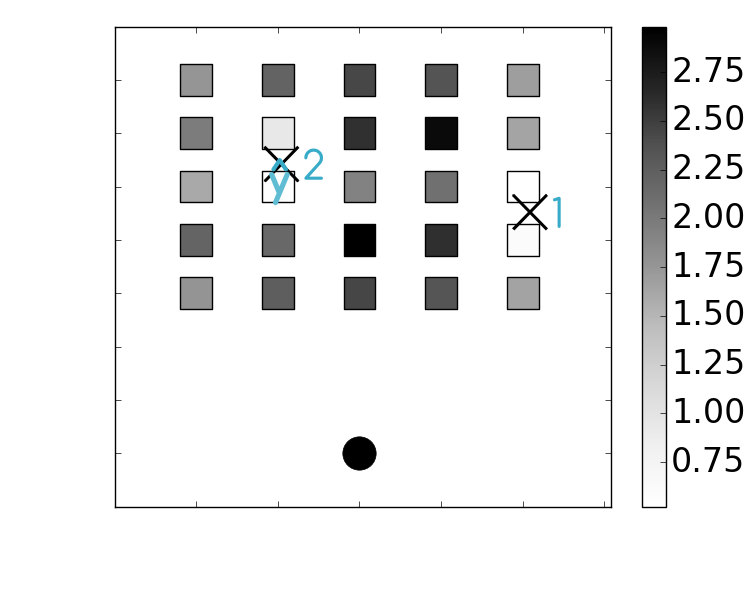}\\[-1mm]
(a) Baseline (2 sensors)
\end{minipage}
\hspace{2mm}
\begin{minipage}[c]{45mm}
 \centering
 \includegraphics[clip,trim=80 70 0 0,height=40mm,width=45mm]{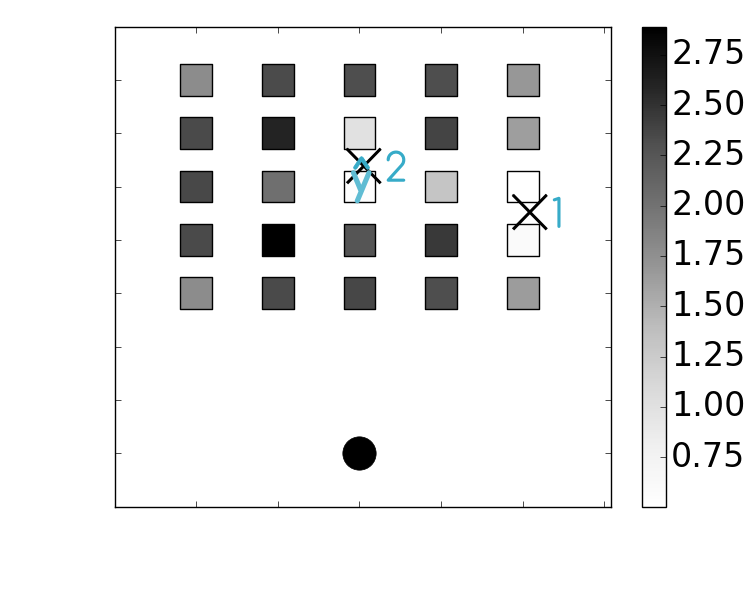}\\[-1mm]
\hspace{-9mm}(b) Proposed (2 sensors)
\end{minipage}
\\[2mm]
%% 5
\begin{minipage}[c]{38mm}
 \centering
 \includegraphics[clip,trim=80 70 80 0,height=40mm,width=38mm]{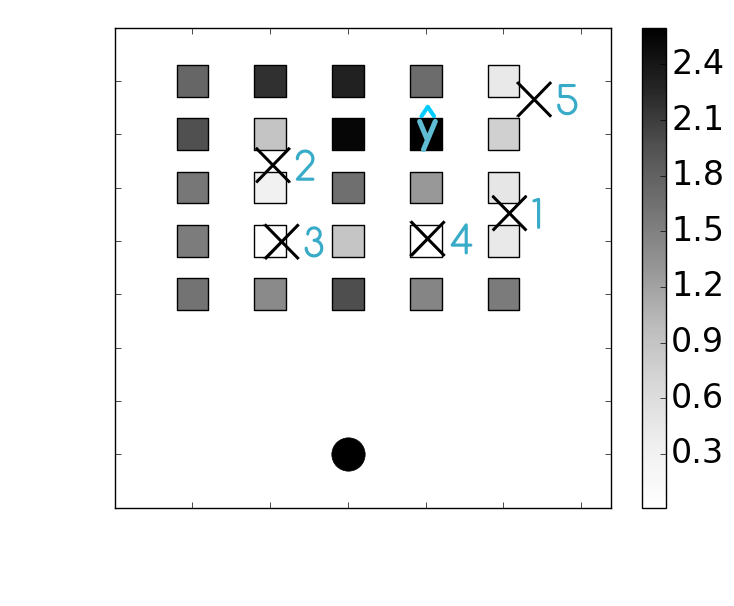}\\[-1mm]
(c) Baseline (5 sensors)
\end{minipage}
\hspace{2mm}
\begin{minipage}[c]{45mm}
 \centering
 \includegraphics[clip,trim=80 70 0 0,height=40mm,width=45mm]{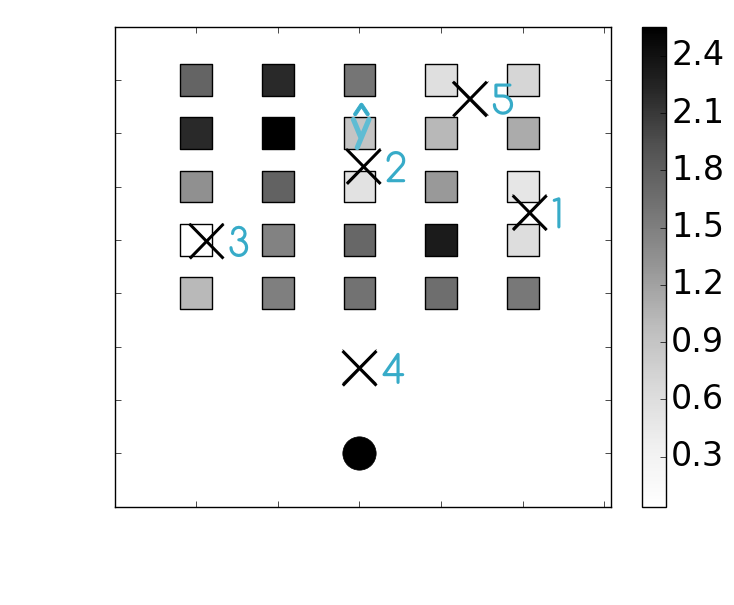}\\[-1mm]
\hspace{-8mm}(d) Proposed (5 sensors)
\end{minipage}
\\[2mm]
%% 8
\begin{minipage}[c]{38mm}
 \centering
 \includegraphics[clip,trim=80 70 80 0,height=40mm,width=38mm]{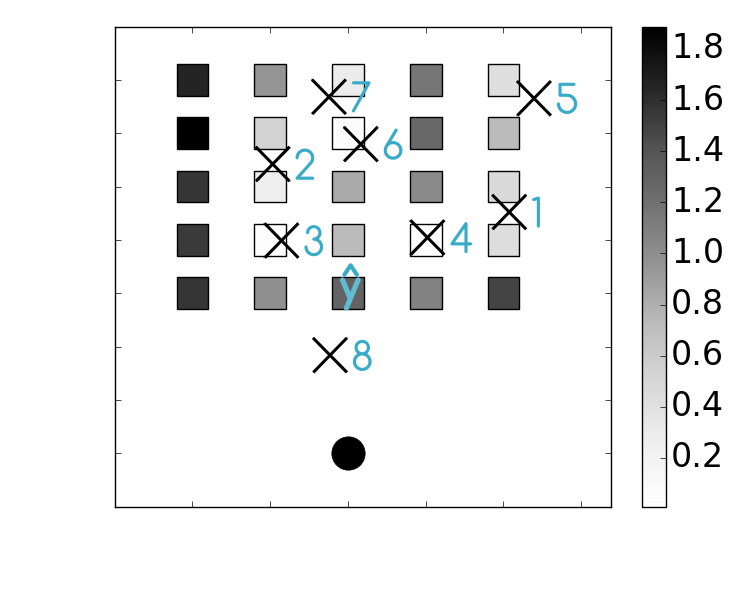}\\[-1mm]
(e) Baseline (8 sensors)
\end{minipage}
\hspace{2mm}
\begin{minipage}[c]{45mm}
 \centering
 \includegraphics[clip,trim=80 70 0 0,height=40mm,width=45mm]{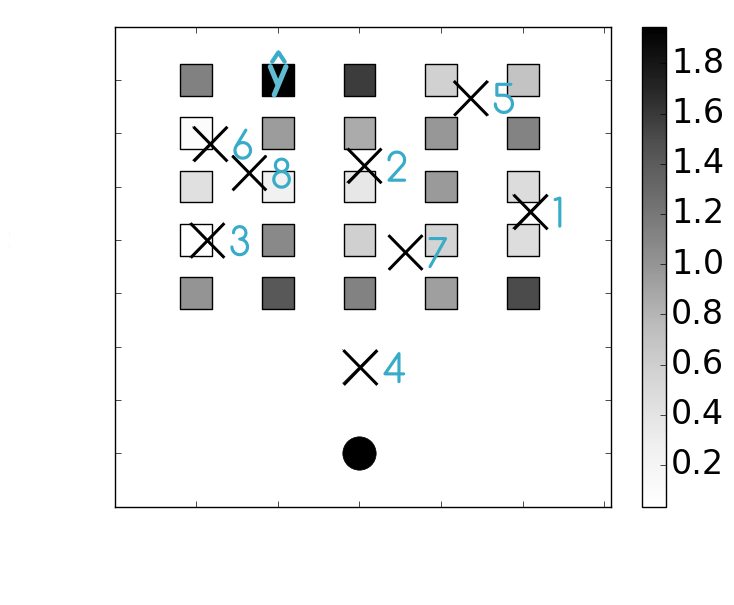}\\[-1mm]
\hspace{-8mm}(f) Proposed (8 sensors)
\end{minipage}
 \caption{
Sensor scattering results by the baseline\cite{K24} (left columns) and SuMo-SS (right columns).
The color strength represents the value of $\delta_y$ after three conditions: (a)(b) 2 sensors, (c)(d) 5 sensors, and (e)(f) 8 sensors. Sub-figure (e) shows the sensors are scattered in a biased way, which is quantitatively validated later in \figref{cumuMIcomp}. The squares and black circles represent $V$ and the loading position, respectively. 
Numbers in blue represents the ordering of $y_{pos}$. In each subfigure, '$\hat{y}$' in blue represents each penultimate target position $\hat{y}_{pos}$.
}
\label{quali_0302}
\end{figure}

% Local Variables:
% eval: (auto-fill-mode -1)
% coding: utf-8
% End:

We compared our proposed method (SuMo-SS) and a baseline method. We used the method proposed by Krause\cite{K24} as the baseline. Unlike SuMo-SS, the baseline method does not consider the uncertainty of sensor positions.

\figref{quali_0302} shows the qualitative results, where $(w_1, w_2)$ are set to $(w_1, w_2)=(0.3,0.2)$. The subfigures in the left and right columns are the results of the baseline and SuMo-SS, respectively. The color strength represents $\delta_y$ \Update (increase in mutual information when sensor $y$ is added) \Done after three conditions: (a) and (b) show two sensors, (c) and (d) show five sensors, and (e) and (f) show eight sensors. In the subfigures, the squares, cross marks (``\ding{53}''), and black circles represent $V$ \Update (set of target position candidates)\Done, $y_{pos}$ \Update (actual position of sensor $y$)\Done, and the loading position, respectively. Note that $y_{pos}$ was unobservable from the methods.
\Update
Numbers in blue represents the ordering of $y_{pos}$. In each subfigure, '$\hat{y}$' in blue represents each penultimate target position $\hat{y}_{pos}$.
\Done

The difference in the sensor positions in subfigures (a) and (b) is thought to be caused by deviation, and this indicates that SuMo-SS and the baseline have no significant difference when two sensors are scattered. Although there is a difference between subfigures (c) and (d), the bias in the sensor positions is not significant. By contrast, subfigure (e) shows that the sensors are scattered in a biased manner compared with subfigure (f). This indicates that SuMo-SS could plan to scatter sensors unbiasedly under uncertainty. Because this needs to be quantitatively validated, we show the quantitative validation in \figref{cumuMIcomp}.

\subsection{Quantitative Results}

\begin{figure}[t]
\begin{minipage}[c]{40mm}
 \centering
 \includegraphics[clip,trim=0 0 30 10,width=40mm,height=38mm]{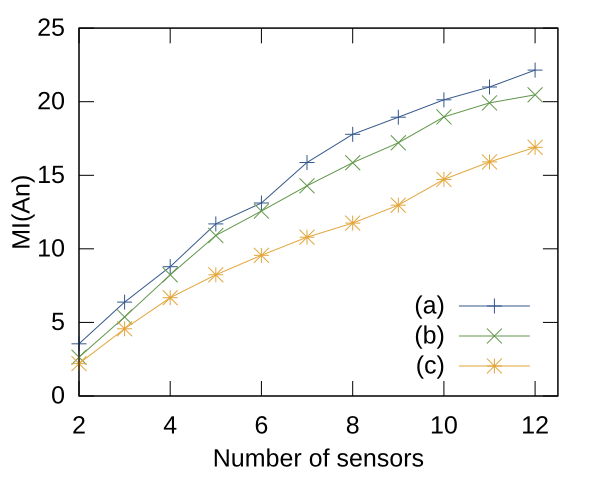}
\end{minipage}
\hspace{1mm}
\begin{minipage}[c]{40mm}
 \centering
 \includegraphics[clip,trim=0 0 30 10,width=40mm,height=38mm]{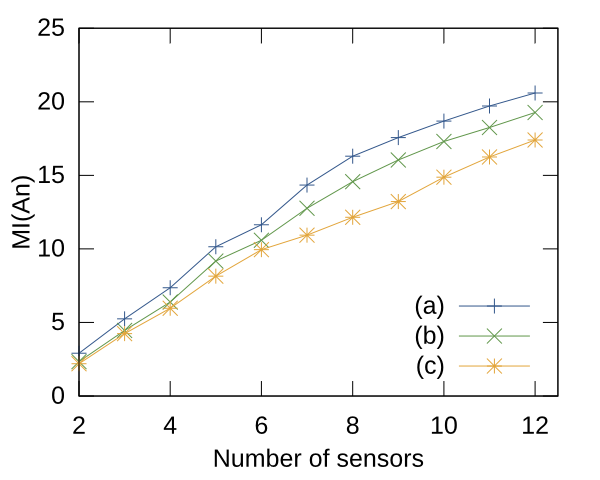}
\end{minipage}
 \caption{
 Comparison of (a) SuMo-SS (proposed), (b) baseline\cite{K24}, and (c) random method.
 $MI(A_n)$ is plotted against the number of sensors, $n$. The average results of ten experimental runs are shown.
 Left: $(w_1, w_2)=(0.3,0.2)$. Right: $(w_1, w_2)=(0.35,0.35)$.
}
\label{cumuMIcomp}
\end{figure}

A quantitative comparison is shown in \figref{cumuMIcomp}. The horizontal axis represents the number of sensors, $n$. The vertical axis represents $MI(A_n)$, which is the mutual information when $n$ sensors are introduced to the environment. $MI(A_n)$ is defined as follows:
\begin{align}
 MI(A_n) = \sum_{i=2}^{n} \delta_{y_i},
\end{align}
where $\delta_{y_i}$ denotes the information gain when the $i$-th sensor is introduced. To satisfy the condition that $MI(A_n)$ is a monotonic submodular function, $n$ needs to be less than $|V|/2$. SuMo-SS and the baseline method require at least one sensor in the environment. Therefore, the first target position was manually given as the center of the area. Then, the second to $n$-th target positions were planned by the proposed and baseline methods.

\figref{cumuMIcomp} compares the results from (a) SuMo-SS (proposed), (b) baseline\cite{K24}, and (c) random selection. The random selection method was introduced as the lower bound method, which selected \Update the next target position \Done $\hat{y}_{pos}$ randomly from $V$. The left-hand and right-hand figures show the results where $(w_1, w_2)=(0.3,0.2)$ and $(w_1, w_2)=(0.35,0.35)$, respectively. The average results of 10 experimental runs are shown.

From the left-hand figure of \figref{cumuMIcomp}, SuMo-SS obtained larger $MI(A_n)$ than the baseline method. $MI(A_n)$ values at $n=12$ obtained by (a) SuMo-SS (proposed), (b) baseline\cite{K24}, and (c) random selection were 22.14, 20.47, and 16.89, respectively. In the right-hand figure, $MI(A_n)$ at $n=12$ obtained by (a), (b), and (c), were 20.59, 19.26, and 17.40, respectively.
The above results show that our method obtained better results in both settings.
% The right-hand figure of \figref{cumuMIcomp} also shows that SuMo-SS outperformed the baseline method.

\subsection{Sensitivity Analysis}

To validate SuMo-SS under various deviations, we evaluated the performance under various combinations of $(w_1, w_2)$. The conditions for $w1$ and $w2$ were $w1 \in \{0.2, 0.25, 0.3, 0.35, 0.4, 0.45, 0.5\}$ and $w2 \in \{0.2, 0.25, 0.3, 0.35, 0.4, 0.45, 0.5\}$, respectively. We ran 10 simulations for each combination of $(w_1, w_2)$. The evaluation was conducted for SuMo-SS and the baseline\cite{K24}. Therefore, we ran the simulation 980 ($=7\times7\times10\times2$) times in total.

\tabref{delta02} shows a performance difference between SuMo-SS and the baseline. The performance difference $\Delta_n$ is defined as follows:
\begin{align}
 \Delta_n = MI(A_n)_{proposed} - MI(A_n)_{baseline},
\end{align}
where $n$ represents the number of sensors. A positive $\Delta_n (> 0)$ indicates that SuMo-SS obtained larger $MI(A_n)$ than the baseline. The subtables (a), (b), (c), and (d) show $\Delta_3$, $\Delta_6$, $\Delta_9$, and $\Delta_{12}$, respectively. 
\Update
In each sub-table, the top and bottom three results are displayed in red and blue, respectively.
\Done

Sub-table (a) shows that SuMo-SS outperformed the baseline in all conditions (49 out of 49) when the number of introduced sensors was three ($n=3$). Sub-tables (b), (c) and (d) show that SuMo-SS outperformed the baseline in 41, 46, and 44 conditions when $n=6$, $n=9$, and $n=12$, respectively. These results indicate that SuMo-SS could obtain larger $MI(A_n)$ under most of the conditions.

\begin{table}[t]
 \centering
 \caption{Performance difference $\Delta_N$ at $N=3, 6, 9, 12$. Average of ten experiments are shown.}
 \label{delta02}
 (a) $N=3$
 \begin{tabular}{|c|c|c|c|c|c|c|c|c|}
% \begin{tabular}{|m{3mm}|m{5mm}|m{5.5mm}|m{5.5mm}|m{5.5mm}|m{5.5mm}|m{5.5mm}|m{5.5mm}|m{5.5mm}|}
\hline 
\multicolumn{2}{|c|}{\multirow{2}{*}{$\Delta_3$}} & \multicolumn{7}{c|}{$w_2$}\tabularnewline
\cline{3-9}
\multicolumn{2}{|c|}{} & 0.2 & 0.25 & 0.3 & 0.35 & 0.4 & 0.45 & 0.5\tabularnewline
\hline 
\multirow{7}{*}{$w_1$} & 0.2 & 1.05 & 1.05 & 0.76 & 0.68 & 0.73 & 0.77 & 0.44 \tabularnewline
\cline{2-9}
 & 0.25 & 1.14 & 1.09 & 0.88 & 0.81 & 0.99 & 0.89 & 0.52 \tabularnewline
\cline{2-9}
 & 0.3 & 1.01 & 0.59 & 0.72 & 0.90 & \textcolor{red}{1.25} & \textcolor{red}{1.20} & \textcolor{red}{1.16} \tabularnewline
\cline{2-9}
 & 0.35 & 0.95 & \textcolor{blue}{0.32} & 0.37 & 0.79 & 1.01 & 0.99 & 0.83 \tabularnewline
\cline{2-9}
 & 0.4 & 1.03 & 0.54 & 0.61 & 0.79 & 0.94 & 0.88 & 0.69 \tabularnewline
\cline{2-9}
 & 0.45 & 0.67 & 0.34 & 0.63 & 0.69 & 0.82 & 0.74 & 0.57 \tabularnewline
\cline{2-9}
 & 0.5 & \textcolor{blue}{0.18} & \textcolor{blue}{0.24} & 0.45 & 0.51 & 0.83 & 0.73 & 0.57 \tabularnewline
\hline
\end{tabular}

 \\[3mm]
 (b) $N=6$
 % \begin{tabular}{|m{3mm}|m{5mm}|m{5.5mm}|m{5.5mm}|m{5.5mm}|m{5.5mm}|m{5.5mm}|m{5.5mm}|m{5.5mm}|}
\begin{tabular}{|c|c|c|c|c|c|c|c|c|}
\hline 
\multicolumn{2}{|c|}{\multirow{2}{*}{$\Delta_6$}} & \multicolumn{7}{c|}{$w_2$}\tabularnewline
\cline{3-9}
\multicolumn{2}{|c|}{} & 0.2 & 0.25 & 0.3 & 0.35 & 0.4 & 0.45 & 0.5\tabularnewline
\hline 
\multirow{7}{*}{$w_1$} & 0.2 & 0.96 & 0.09 & 0.92 & \textcolor{red}{1.35} & \textcolor{red}{1.34} & \textcolor{red}{1.22} & 0.57 \tabularnewline
\cline{2-9}
 & 0.25 & 0.84 & 0.66 & 0.95 & 0.86 & 0.85 & 1.03 & 0.44 \tabularnewline
\cline{2-9}
 & 0.3 & 0.55 & -0.21 & 0.59 & 0.65 & 0.66 & 0.25 & 0.83 \tabularnewline
\cline{2-9}
 & 0.35 & 0.45 & -0.15 & 0.77 & 1.05 & 0.80 & 0.77 & 0.68 \tabularnewline
\cline{2-9}
 & 0.4 & 0.46 & 0.05 & 1.02 & 0.85 & 0.94 & 0.94 & 1.13 \tabularnewline
\cline{2-9}
 & 0.45 & -0.39 & -0.21 & 0.72 & 0.32 & 0.73 & 0.55 & 0.64 \tabularnewline
\cline{2-9}
 & 0.5 & \textcolor{blue}{-0.82} & \textcolor{blue}{-1.17} & -0.16 & \textcolor{blue}{-0.42} & 0.22 & 0.03 & 0.32 \tabularnewline
\hline
\end{tabular}

 \\[3mm]
 (c) $N=9$
 % \begin{tabular}{|m{3mm}|m{5mm}|m{5.5mm}|m{5.5mm}|m{5.5mm}|m{5.5mm}|m{5.5mm}|m{5.5mm}|m{5.5mm}|}
\begin{tabular}{|c|c|c|c|c|c|c|c|c|}
\hline 
\multicolumn{2}{|c|}{\multirow{2}{*}{$\Delta_9$}} & \multicolumn{7}{c|}{$w_2$}\tabularnewline
\cline{3-9}
\multicolumn{2}{|c|}{} & 0.2 & 0.25 & 0.3 & 0.35 & 0.4 & 0.45 & 0.5\tabularnewline
\hline 
\multirow{7}{*}{$w_1$} & 0.2 & 0.73 & \textcolor{blue}{-0.28} & 0.29 & 1.86 & 0.80 & 1.54 & 0.37 \tabularnewline
\cline{2-9}
 & 0.25 & 0.76 & 0.23 & 1.34 & 1.99 & 0.73 & 1.46 & 0.22 \tabularnewline
\cline{2-9}
 & 0.3 & 1.73 & \textcolor{blue}{-0.11} & 1.31 & \textcolor{red}{2.08} & 1.08 & 0.45 & 1.43 \tabularnewline
\cline{2-9}
 & 0.35 & 1.96 & \textcolor{blue}{-0.12} & 1.46 & 1.52 & 1.00 & 0.30 & 0.51 \tabularnewline
\cline{2-9}
 & 0.4 & \textcolor{red}{2.06} & 0.03 & \textcolor{red}{2.32} & 1.50 & 0.72 & 0.68 & 1.06 \tabularnewline
\cline{2-9}
 & 0.45 & 1.03 & 0.08 & 0.89 & 0.55 & 0.81 & 0.60 & 1.26 \tabularnewline
\cline{2-9}
 & 0.5 & 0.29 & 0.16 & 0.30 & 0.47 & 0.12 & 0.27 & 0.69 \tabularnewline
\hline
\end{tabular}

 \\[3mm]
 (d) $N=12$
 % \begin{tabular}{|p{3mm}|p{4.5mm}|p{5.7mm}|p{5.7mm}|p{5.7mm}|p{5.7mm}|p{5.7mm}|p{5.7mm}|p{5.7mm}|}
\begin{tabular}{|c|c|c|c|c|c|c|c|c|}
\hline 
\multicolumn{2}{|@{}c@{}|}{\multirow{2}{*}{$\Delta_{12}$}} & \multicolumn{7}{@{}c@{}|}{$w_2$}\tabularnewline
% \multicolumn{2}{|c|}{\multirow{2}{*}{$\Delta_{12}$}} & \multicolumn{7}{c|}{$w_2$}\tabularnewline
\cline{3-9}
% \multicolumn{2}{|c|}{} & 0.2 & 0.25 & 0.3 & 0.35 & 0.4 & 0.45 & 0.5\tabularnewline
\multicolumn{2}{|@{}c@{}|}{} & 0.2 & 0.25 & 0.3 & 0.35 & 0.4 & 0.45 & 0.5\tabularnewline
\hline 
\multirow{7}{*}{$w_1$} & 0.2 & 0.05 & \textcolor{blue}{-0.68} & 0.31 & 1.67 & 0.58 & 0.90 & 0.10 \tabularnewline
\cline{2-9}
 & 0.25 & 0.81 & 0.22 & 0.55 & 1.27 & 0.43 & 0.45 & -0.48 \tabularnewline
\cline{2-9}
 & 0.3 & 1.68 & \textcolor{blue}{-0.84} & 1.55 & 1.61 & 1.50 & 0.29 & 1.59 \tabularnewline
\cline{2-9}
 & 0.35 & \textcolor{red}{1.96} & \textcolor{blue}{-0.87} & \textcolor{red}{2.00} & 1.33 & 1.59 & 0.35 & 1.00 \tabularnewline
\cline{2-9}
 & 0.4 & 1.55 & -0.05 & \textcolor{red}{2.94} & 1.41 & 0.73 & 0.68 & 1.19 \tabularnewline
\cline{2-9}
 & 0.45 & 0.73 & 0.15 & 1.58 & 0.70 & 1.10 & 0.51 & 1.45 \tabularnewline
\cline{2-9}
 & 0.5 & 0.13 & -0.06 & 0.41 & 0.13 & 0.34 & 0.15 & 0.78 \tabularnewline
\hline
\end{tabular}

\end{table}

% Local Variables:
% eval: (auto-fill-mode -1)
% coding: utf-8
% End:

\subsection{Discussions
}

First, we discuss the covariance of the sensor observations. In this study, covariance was obtained based on the sensor positions. 
\Update
This does not mean that SuMo-SS requires precise sensor position information.
\Done
Instead, this was because (a) our focus is not to model realistic sensor observations, and (b) simulations require a certain-level of approximation on sensor observation. However, this does not hold in real-world applications; therefore, covariance should be calculated based on sensor observations. By doing so, we will be able to apply the proposed method to real-world applications including cases in which scattered sensors are washed away by rain.

We used mutual information $MI(A)$ as the criterion for submodular optimization. However, we can use other criteria that have submodularity, such as the monitoring area size and the number of grid points covered by the area.
\Update
Future study includes the improvement of the optimization policy instead of the greedy method. Golovin et al. proposed adaptive policies by introducing the concept of adaptive submodularity\cite{G8}. Although the assumptions for submodularity and adaptive submodularity are different, there is a possibility that the SS problem can be extended to satisfy the assumptions.
\Done

\Update
\figref{cumuMIcomp} might give the impression that the performance of the baseline and SuMo-SS slightly decrease at $n=12$. This is caused by the fact that the increase in $\delta_{y_i}$ is decreasing in monotonic submodular functions. \Done In this study, the maximum number of sensors was 12, which is the greatest integer less than $|V|/2$. However, this does not mean that the method is limited to 12 sensors. By increasing $|V|$, more sensors can be deployed without any fundamental changes. \Update For example, if a developer needs to deploy 100 sensors in a practical use case, then $|V|$ can be set $|V|=201, 202, ...$ because $|V|$ is arbitrary in SuMo-SS. \Done

% In \figref{cumuMIcomp}, the maximum number of sensors was 12, which is the greatest integer less than $|V|/2$. However, this does not mean that the method is limited to 12 sensors. By increasing $|V|$, more sensors can be deployed without any fundamental changes.

One might question whether sensors should be simply dropped at grid points; however, not all sensors might be informative because sometimes local events cannot be monitored by rough granularity. Although the sensor material can be changed to reduce the deviation, reducing it to zero will not be easy. Although we used a drone to transport sensors, SuMo-SS can be applied to a setting in which sensors are deployed with catapult-like devices provided that the deviation can be modeled.

% \begin{itemize}
%  \item In this paper, we consider the problem where a drone drops a limited number of sensors. Because battery life is limited, optimal sensor placement is desirable.
% \end{itemize}

% Local Variables:
% eval: (auto-fill-mode -1)
% coding: utf-8
% End:

\section{Summary}

% Drones have a huge potential impact on a variety of use cases for monitoring security situations, agriculture, amusement, and so on. According to an investigation by the Association for Unmanned Vehicle Systems International (AUVSI), the economic impact by drones in the US is estimated to be approximately 82.1 billion dollars during the period from 2015 to 2025\cite{J7}.

In this paper, we made the following contribution:
\begin{itemize}
 % \item We introduced submodular optimization to drone-based sensor deployment. 
 %       The proposed method does not suffer from combinatorial explosion and obtains $(1-1/e)$--approximation.
 \item We proposed the SuMo-SS method that can deal with uncertainty in sensor positions. Unlike existing methods, SuMo-SS can deal with uncertainty in sensor positions, which is relevant for practical applications. Its experimental validation with a baseline method was explained in \secref{exp}.
 % \item We proposed an SS method by maximizing expected mutual information.
 % \item We validated our method by comparing it with the baseline methods in simulation experiments.
\end{itemize}
The target use case of our method include building sensor networks for environmental monitoring.
Future work includes an experimental validation with physical drones in outdoor environments.

% Local Variables:
% eval: (auto-fill-mode -1)
% coding: utf-8
% End:

\appendix
\section{Appendix}

\subsection{Submodurality in Expected Mutual Information}

\Update
A set function $f$ is called submodular if it holds $f(A \cup \{e\}) - f(A) \geq f(B \cup \{e\}) - f(B)$
for every $A, B \subseteq E$ with $A \subseteq B$ and every $e \in E \backslash B$. $MI(A)$ is proved to be a monotone submodular function in particular conditions\cite{N10,K24}.

If the probabilistic distribution of $y_{pos}$ is discrete, \eqref{eq43} can be rewritten as follows:
\begin{align}
 \hat{y}_{pos} &= 
\argmax_{y \in V\backslash A} 
% \mathbb{E}_{pos}[MI(A\cup y) - MI(A)].
\sum p(y_{pos})[MI(A\cup y) - MI(A)],
\label{eqA7}
\end{align}
where $p(y_{pos})$ denotes the probabilistic distribution of $y_{pos}$. A nonnegative linear combination of submodular functions is also submodular\cite{K28,K29}. Therefore, the right-hand term in \eqref{eqA7} is submodular. If $p(y_{pos})$ is continuous instead, we can approximate it with the average of sufficiently fine discrete distributions as shown in \eqref{eqA7}. Therefore, the expected mutual information shown in \eqref{eq43} is a submodular function.
\Done

\subsection{Submodular Optimization Using Mutual Information}

Hereinafter, we explain the method proposed in \cite{K24}. For readability, $y_{obs}$ and $\mathcal{Y}_A$ are written as $y$ and $A$, respectively.

From the definition of mutual information, $MI(A)$ is decomposed as follows:
\begin{align*}
 MI(A)
 &= H(A) - H(A|V \backslash A) =  H(A) - H(A| \bar{A} \cup y )\\
 MI(A \cup y) &= H(A \cup y) - H(A \cup y | \bar{A}),
\end{align*}
where $H(\cdot)$ represents entropy.

Let $\delta_y$ be the difference between $MI(A \cup y)$ and $MI(A)$ as follows:
% The difference of $MI(A \cup y) - MI(A)$ is represented as $\delta_y$, which can be transformed as follows:
\begin{align}
 \delta_y &= MI(A \cup y) - MI(A) \nonumber\\
 &= H(A \cup y) - H(A \cup y | \bar{A}) - H(A) + H(A|\bar{A} \cup y) \label{eqA2}
.
\end{align}
From the definition of conditional entropy, $H(A \cup y | \bar{A}) \}$ can be written:
\begin{align}
 H(A \cup y | \bar{A}) \} &= H(A \cup y, \bar{A}) - H(\bar{A}) \nonumber \\
 &= H(V) - H(\bar{A}).
\end{align}
We can also transform $H(A|\bar{A} \cup y)$ in the same manner. Thus \eqref{eqA2} can be rewritten:
\begin{align}
 \delta_y
 % &= H(A \cup y) - H(V) + H(\bar{A}) - H(A) + H(V) - H(\bar{A} \cup y) \nonumber\\
 &= H(A \cup y) - H(A) - H(\bar{A} \cup y) + H(\bar{A}) \nonumber\\
 &= H(y| A) - H(y|\bar{A}) . \label{eqA3}
\end{align}

Another definition of conditional entropy $H(y|A)$ is given as follows:
\begin{align}
 H(y|A)
 &= -\int p(y,A) \log \mathcal{N} (\mu_{y|A}, \sigma^2_{y|A})dydA \nonumber\\
 % &= -\int p(y,A) \log \mathcal{N} (\mu_{y|A}, \sigma^2_{y|A})dydA\\
 &= \frac{1}{2} \log 2 \pi e \sigma^2_{y|A}, \label{eqA4}
\end{align}
where the formula of the integral of Gaussian distributions is used. Similarly, we can obtain $H(y|\bar{A})$. 
From Equations (\ref{eqA3}) and (\ref{eqA4}), we obtain
\begin{align}
 \delta_y = \frac{1}{2} \log \frac{\sigma^2_{y|A}}{\sigma^2_{y|\bar{A}}}. \label{eqA5}
\end{align}

When a multi-variate Gaussian distribution is divided, the following holds:
\begin{align}
 % \mu_{y|A} &= \mu_y + \vec{\Sigma}_{yA} \vec{\Sigma}_{AA}^{-1} ( \mathcal{Y}_A - {\bm \mu}_A), \\
 \sigma^2_{y|A} &= \sigma^2_y - \vec{\Sigma}_{yA} \vec{\Sigma}_{AA}^{-1} \vec{\Sigma}_{Ay}. \label{eqA51}
\end{align}

% Finally, we obtain the following:
From Equations (\ref{eqA5}) and (\ref{eqA51}), we obtain the following:
\begin{align}
 \delta_y = \frac{1}{2} \log
\frac{ \sigma^2_y - \vec{\Sigma}_{yA} \vec{\Sigma}_{AA}^{-1} \vec{\Sigma}_{Ay} } 
{\sigma^2_y - \vec{\Sigma}_{y\bar{A}} \vec{\Sigma}_{\bar{A}\bar{A}}^{-1} \vec{\Sigma}_{\bar{A}y}}.
 \label{eqA6}
\end{align}

% Local Variables:
% coding: utf-8
% End:

\section*{ACKNOWLEDGMENT}

This work was partially supported by JST CREST and JSPS KAKENHI Grant Number JP15K16074.

\bibliographystyle{IEEEtran}
\bibliography{reference}

\end{document}